\documentclass[letterpaper, 10 pt, journal, twoside]{IEEEtran} 
\usepackage{amsmath,amsfonts}
\usepackage{algorithmic}
\usepackage{algorithm}
\usepackage{array}
\usepackage{textcomp}
\usepackage{url}
\usepackage{verbatim}
\usepackage{graphicx}
\usepackage{cite}
\usepackage{graphics} 
\usepackage{amssymb}  
\usepackage{subcaption}
\usepackage{xcolor}
\usepackage{booktabs}
\usepackage{multirow}
\usepackage{float}
\usepackage{hyperref}
\hypersetup{
    colorlinks=false,
    citebordercolor=white,
    linkbordercolor=white,
    urlbordercolor=magenta,
    }

\usepackage{pifont}
\newcommand{\cmark}{\ding{51}}%
\newcommand{\xmark}{\ding{55}}%

\newcommand{\alignedtriangle}{{\raisebox{0.15ex}{\scalebox{1.0}{{\color{red}$\blacktriangle$}}}}}

\newcommand{\alignedstar}{{\raisebox{0.15ex}{\scalebox{0.9}{\color{green!50!black}$\bigstar$}}}}

\newcommand{\alignedbullet}{{\raisebox{-0.15ex}{\scalebox{1.2}{\color{blue}$\bullet$}}}}

\DeclareMathOperator*{\argmin}{arg\,min}

\newcommand{\bftab}{\fontseries{b}\selectfont}

\newcommand{\blueup}{\textcolor{blue}{$\uparrow$}}
\newcommand{\greendown}{\textcolor{green}{$\downarrow$}}

\begin{document}

\title{GISR: Geometric Initialization and Silhouette-based Refinement for Single-View Robot Pose and Configuration Estimation}

\author{Ivan Bilić$^{1}$, Filip Marić$^{1,2}$, Fabio Bonsignorio$^{1}$, Ivan Petrović$^{1}$%
\thanks{Manuscript received: May 7, 2024; Revised August 5, 2024; Accepted September 9, 2024.}
\thanks{This paper was recommended for publication by Editor Markus Vincze upon evaluation of the Associate Editor and Reviewers' comments. 
This work has been supported by the H2020 project AIFORS under Grant Agreement No 952275.} 
\thanks{$^{1}$Authors are with University of Zagreb Faculty of Electrical Engineering and Computing, Laboratory for Autonomous Systems and Mobile Robotics, Croatia
        {\tt\footnotesize ivan.bilic@fer.hr}}%
\thanks{$^{2}$Filip Marić is with University of Toronto Institute for Aerospace Studies, Space and Terrestrial Autonomous Robotic Systems Laboratory, Canada
        {\tt\footnotesize filip.maric@mail.utoronto.ca}}%
\thanks{Digital Object Identifier (DOI): see top of this page.}
}

\markboth{IEEE Robotics and Automation Letters. Preprint Version. Accepted September, 2024}
{Bilić \MakeLowercase{\textit{et al.}}: Geometric Initialization and Silhouette-based Refinement for Single-View Robot Pose and Configuration Estimation}


\maketitle

\begin{abstract}
In autonomous robotics, measurement of the robot’s internal state and perception of its environment, including interaction with other
agents such as collaborative robots, are essential. 
Estimating the pose of the robot arm from a single view has the potential to replace classical eye-to-hand calibration approaches and is particularly attractive for online estimation and dynamic environments.
In addition to its pose, recovering the robot configuration provides a complete spatial understanding of the observed robot that can be used to anticipate the actions of other agents in advanced robotics use cases.
Furthermore, this additional redundancy enables the planning and execution of recovery protocols in case of sensor failures or external disturbances. 
We introduce GISR - a deep configuration and robot-to-camera pose
estimation method that prioritizes execution in real-time.
GISR consists of two modules: (i) a geometric initialization module that efficiently computes an approximate robot pose and configuration, and (ii) a deep iterative silhouette-based refinement module that arrives at a final solution in just a few iterations.
We evaluate GISR on publicly available data and show that it outperforms existing methods of the same class in terms of both speed and accuracy, and can compete with approaches that rely on ground-truth proprioception and recover only the pose. Our code is available at \href{https://github.com/iwhitey/GISR-robot}{https://github.com/iwhitey/GISR-robot}.
\end{abstract}


\section{Introduction}
\IEEEPARstart{R}{obotics} applications in a wide array of domains rely on accurate camera-to-robot pose and configuration measurements.
In applications such as robotic grasping, finding the camera-to-robot pose is crucial, as sensor-centered measurements of interest (e.g., object detections) need to be transformed to the robot's reference frame \cite{tremblay2020indirect}.
Once spatial analysis in the context of the task is complete, the robot is set in motion using control algorithms that rely on accurate proprioceptive sensing from joint encoders.
However, robots that operate in hazardous environments (e.g., nuclear decommissioning) or perform strenuous tasks (e.g., heavy loading) often lack proprioceptive sensors \cite{vision_based_control}.
Similarly, failure in proprioceptive sensing may be detrimental in applications where robots perform tasks in environments not easily accessible to humans (e.g., underwater welding~\cite{underwater_welding}), incurring high operational delay and maintenance expenses.
Redundancy offered by additional proprioceptive estimates allows for the design of recovery protocols that could help avoid many consequences of system failure.
Moreover, the ability to jointly estimate the robot pose and configuration using visual cues can be of great value in scenarios with dynamic and unstructured environments, improving collaboration between robots~\cite{cooperative_manipulation} and enhancing safety in human-robot collaboration~\cite{human_robot_collab}.

\begin{figure}[t]
\centering
\includegraphics[width=0.4\textwidth]{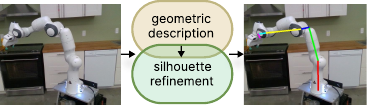}
\caption{GISR takes an input RGB image of the robot (left) and outputs an estimate of both the camera-to-robot pose and the configuration of the robot. The corresponding skeleton is projected onto the input image and overlaid (right).}
\label{fig:goal_task}
\end{figure}

We consider the estimation of both the 6D camera-to-robot pose and the robot's joint angle values (i.e., configuration), using a single RGB image as input, as shown in Fig.~\ref{fig:goal_task}.
The use of off-the-shelf cameras for image-based estimation of robot configurations is cost-effective in terms of hardware, provides a non-invasive way to monitor robot configurations and offers a large degree of flexibility in sensor placement.
Recent works have proposed methods for estimating either the camera-to-robot pose or robot configuration from a single RGB image based on keypoint detection~\cite{dream, lu2023markerless, craves}, rendering-based refinement \cite{robopose}, and geometry-aware models \cite{bilic_maric_ifac_2023}, demonstrating favorable performance compared to classical marker-based approaches.
%
%
For example, keypoint-based methods \cite{dream, lu2023markerless, craves, bilic_maric_ifac_2023} can recover the 6D pose or configuration in real-time, but require ground-truth configuration or exhibit very limited accuracy.
Conversely, rendering-based methods \cite{robopose} can recover both, but their dense and iterative nature incurs a high computational cost to do so very accurately.

In this paper, we present GISR - a method that can recover camera-to-robot-pose and configuration estimates in real-time using a strong geometric prior and dense silhouette-based refinement.
GISR leverages the expressive power of Deep Neural Networks (DNNs) and a geometric prior associated with the robot's kinematic model to realize two modules that complement each other.
%
The geometric initialization module (GIM) produces an initial estimate of the camera-to-robot pose and robot configuration using the distance-geometric model introduced in~\cite{TRO_Maric, giamou2022convex}.
Using this initial estimate, the refinement module (RM) generates a corresponding silhouette image, which is used alongside the segmented input image to predict an update.
Assuming known camera intrinsics, we posit that a rendered silhouette of the robot contains most information essential for the joint estimation task.
Moreover, excluding RGB image details like background and lighting helps avoid the fidelity gap due to rendering imperfections and environment variations.
Therefore, using silhouette images simplifies learning by focusing only on critical information, offering faster rendering and reduced fidelity discrepancies.

Unlike similar methods that focus on only one aspect, GISR is able to recover both the camera-to-robot pose and joint configuration.
Our implementation of GISR exhibits a running time in the order of $40$ms, which is $20 \times$ faster than existing dense methods and comparable to the speed of keypoint-based methods that rely on ground-truth configuration measurements.
We provide a quantitative and qualitative analysis of our approach on a high-DoF robot using a publicly available dataset.

\section{RELATED WORK}
\label{sec:related_work}

\subsection{Object pose estimation}
Estimating the 6D pose of rigid objects from a single RGB image is a long-standing goal in computer vision.
Most relevant to our work are pose estimation approaches based on keypoint detection and refinement.
Within the keypoint-based approach, state-of-the-art methods \cite{pose_from_semantic_keypoints, real_time_6d_pose, tremblay2018deep} train DNNs to detect object features in an image and establish a sparse set of 2D-3D correspondences. 
The camera pose is then recovered using the Perspective-n-Point (PnP) algorithm \cite{epnp} acting on the correspondences.
These methods are usually fast but less robust as they determine the pose from a sparse data.
Further, various methods use a refinement-based approach that iteratively improves the pose prediction by directly regressing the pose updates \cite{deepim, manhardt2018deep}.
These methods are typically more robust, but also more computationally intensive compared to those using the keypoint-based approach.
Overall, a robot manipulator can be seen as an object with multiple degrees of freedom. Therefore, our refinement module is inspired by methods that are trained to iteratively regress object pose updates \cite{robopose, deepim, manhardt2018deep}.

\subsection{Camera-to-robot pose estimation}
Camera-to-robot pose estimation is essential for many robotic applications, e.g., collaborative robotics \cite{cooperative_manipulation}, augmented reality \cite{lambrecht2019markerless} and robotic surgery \cite{robotic_surgery}, to name a few.
The classical approach to determine the camera-to-robot pose, known as hand-eye calibration \cite{handeye}, is based on the use of multiple frames and fiducial markers \cite{Olson2011AprilTagAR}.
More recently, single-view RGB methods, largely influenced by ideas from object pose estimation, have started gaining traction in the robotics research community.
In \cite{dream, lambrecht2019markerless}, DNNs are trained to predict a set of 2D keypoints associated the robot's joints, while their 3D correspondences are found by applying forward kinematics using the known current configuration of the robot.
At test time, the PnP algorithm is then applied to solve for the pose using the established 2D-3D correspondences.
In \cite{lu2023markerless}, the PnP algorithm is integrated into the learning process to train a keypoint detector in a self-supervised manner.
%
However, keypoint-based methods are less robust to viewpoint changes \cite{robopose} and can be affected by a suboptimal specification of keypoint locations \cite{lu2022pose}.
%
In the context of soft robotics, \cite{lu2023image} use a differentiable renderer to create a richer, spatially-informed cost function for an optimization-based alignment of geometric primitives.
Similarly, the authors of \cite{robopose} employ rendering-based refinement detached from the learning process for joint camera-to-robot pose and configuration estimation, resulting in a model that is more robust compared to keypoint-based methods.
Inspired by these approaches, GISR uses a deep refinement module preceded by a geometric module based on~\cite{bilic_maric_ifac_2023}, which learns to generate initial estimates by minimizing a loss that jointly encodes the robot configuration and kinematic structure.
%
%

\begin{figure*}[t]
    \vspace*{5pt}
    \centering
    \centerline{\includegraphics[width=\textwidth]{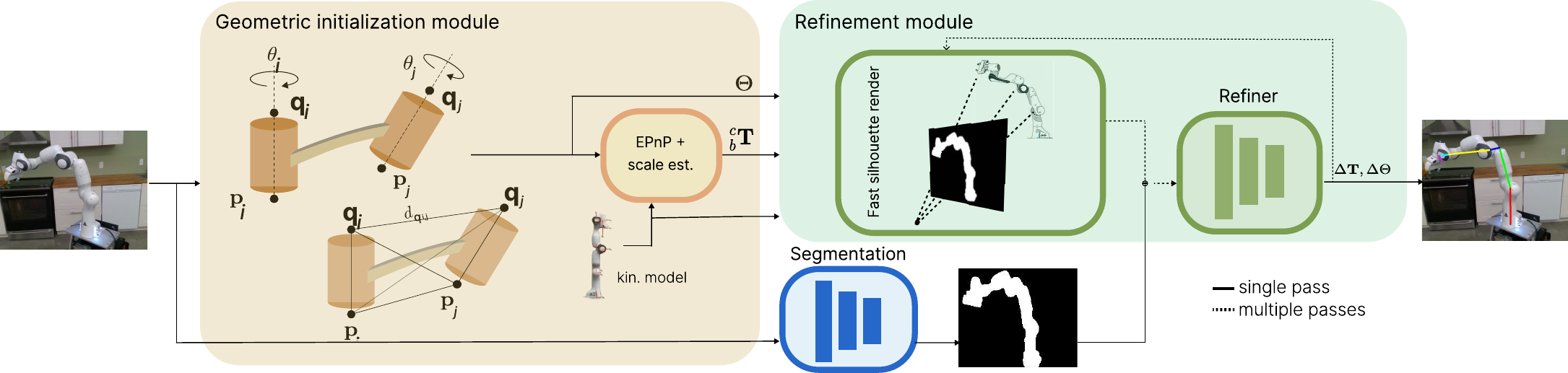}} 
    \caption{System overview. The geometric initialization module\protect\footnotemark (GIM) takes an input RGB image and produces initial estimates of the robot pose and configuration. The refinement module (RM) uses these estimates to generate a corresponding silhouette image, which is fed to the refiner along with the segmented input image to predict an update. This render-and-update process can be repeated, but each iteration requires a forward pass of a deep model (including the update and rendering).}
    \label{fig:system}
\end{figure*}
\footnotetext{The subfigure showing the neighbouring joints is from our previous work \cite{bilic_maric_ifac_2023}.}

\subsection{Joint camera-to-robot pose and configuration estimation}
Jointly estimating the 6D camera-to-robot pose and the robot configuration offers a wider range of applications compared to pose-only or configuration-only estimation, at the cost of greater complexity due to a lack of ground-truth spatial information.
Learning to estimate both quantities may also help increase overall robustness by making the model rely on structural information such as link geometry, while carrying a potentially higher risk of overfitting when the data is insufficient or badly distributed.
Recently, several frameworks have been proposed that solve this or similar problems using only a single RGB image as an input.
In \cite{craves}, the authors propose to train a keypoint detector followed by local optimization, detached from the learning process, to directly regress the full state of a low-cost 4-DoF manipulator.
%
%
Similarly, Labb\'e et. al~\cite{robopose} propose a method inspired by DeepIM \cite{deepim}, where rendered RGB images of the robot model, at an initially random or fixed pose and configuration, are used as inputs to a deep refinement model trained to correct the error, repeating the process iteratively.
Despite this approach exhibiting stronger robustness than keypoint-based methods, using such uninformed initialization strategy combined with per-iteration RGB rendering makes it an order of magnitude slower.
%
%
%

In contrast, the GISR refinement module uses a rendered silhouette of the robot at the previous configuration estimate and the segmented input image to produce an update minimizing the 3D joint location error.
%
In this way, the refinement module learns the distribution of initial estimates produced by the trainable geometric module, reducing the discrepancy between initialization applied at train and test times.
%
%
The silhouette images are generated using a fast rendering procedure that avoids rasterization and shading, which results in a total running time almost 20 times lower than~\cite{robopose}.
%
%
%
%

%
%
%
%
%
\section{PROPOSED METHOD}
\label{sec:methodology}
Given an input RGB image of the robot manipulator comprised of a series of rigid links and revolute joints, our goal is to recover the observed robots' configuration $\mathbf{q} \in \mathcal{C}$, where $\mathcal{C} \subseteq \mathbb{R}^n$ represents the configuration space, and a 6D camera-to-robot pose ${}_{b}^{c}\mathbf{T}$ consisting of a rotation $\mathbf{R} \in \mathit{SO}(3)$ and a translation $\mathbf{t} \in \mathbb{R}^{3}$ with respect to the camera coordinate frame.
As shown in Fig.~\ref{fig:system}, GISR consists of a geometric initialization module and a refinement module, which are described in detail throughout this section.
%
%

\subsection{Geometric initialization module}
\label{subsec:geom_init}
The proposed geometric initialization method enables a fast computation of an informed initial guess of the robot pose $\mathbf{T}$ and the configuration $\mathbf{\Theta}$ observed in an image $\mathcal{I}$.
%
To achieve this, our initialization procedure exploits insights from recent work on distance-based inverse kinematics \cite{bilic_maric_ifac_2023, TRO_Maric}, which show that the configuration of a robot whose kinematic model $\mathbf{\Theta} = \{ \theta_i\}_{i=0}^{d}$ is known can be recovered from spatial constraints arising from a set of 3D points $\mathbf{X} = \{\mathbf{P, Q}\}^{n}$ defined with respect to the robot's kinematic chain.
More specifically, these points are given by:

\begin{equation}
\begin{aligned}
\mathbf{p}_i = C \prod _{j=1}^{i}{}_{j}^{j-1}\mathbf{T}(\theta _{j}) \\ 
\mathbf{q}_i = \mathbf{p}_i + \mathbf{R}_i \hat{\mathbf{z}},
\end{aligned}
\label{eq:points}
\end{equation}
where the 3D joint locations $\mathbf{P}$ are extracted by applying a selection matrix $C$ to the computed pose transformations for the coordinate frame $i \in \{1,...,d\}$ of each link, while the points $\mathbf{Q}$ are defined to be a unit distance away from points $\mathbf{P}$ along the joints’ rotation axes $\hat{\mathbf{z}}$ using their respective orientation $\mathbf{R}_i$.
Instead of recovering the points directly, the task of the geometric module is formulated as recovering the Euclidean Distance Matrix (EDM) that these points generate:

\begin{equation}
\mathbf{D^*} = \argmin_\mathbf{D} \| \hat{\mathbf{D}}(\mathcal{K}, \zeta) - \mathbf{D(\Theta^*)}\|_F,
\label{eq:geometric_goal}
\end{equation}
where $\|\cdot\|_F$ represents Frobenius norm, $\mathcal{K}$ denotes a set of 2D keypoints representing the coordinate frame positions of the joints, as defined by the kinematic model of the robot,
and $\zeta$ is parametrized by a shallow feedforward neural network. $\mathbf{D(\Theta^*)}$ represents a target distance matrix constructed from points defined in (\ref{eq:points}), which uniquely corresponds to the ground-truth configuration $\mathbf{\Theta^*}$ and serves as a geometric description of the robot configuration observed in the image.
Then, the Gram matrix $\mathbf{G}$ corresponding to the estimated $\hat{\mathbf{D}}$ is obtained by taking:
\begin{equation}
\mathbf{G} = -\frac{1}{2}\mathbf{J}\mathbf{\hat{\mathbf{D}}}\mathbf{J},
\label{eq:gram_from_edm}
\end{equation}
where $\mathbf{J} = \mathbf{I} - \frac{1}{n}\mathbf{1}\mathbf{1^{T}}$ is a geometric centering matrix formed by using an identity matrix $\mathbf{I}$ and a column vector of ones $\mathbf{1}$. Finally, since $\mathbf{G} = \mathbf{X}^{T} \mathbf{X}$ is a real symmetric matrix, a set of geometrically centered points $\mathbf{X}$ that generates $\hat{\mathbf{D}}$ is obtained via eigendecomposition \cite{dokmanic}.
The observed configuration is finally recovered by applying a set of kinematic transformations designed to map a set of points $\mathbf{X}$ to a configuration $\mathbf{\Theta}$ analytically, as described in \cite{TRO_Maric}.
Together with (\ref{eq:geometric_goal}) and (\ref{eq:gram_from_edm}), these differentiable transformations form an end-to-end trainable geometric pipeline that predicts an initial estimate for the observed configuration.

To estimate the initial 6D robot-to-camera pose ${}_{b}^{c}\mathbf{T}$, we first use (\ref{eq:points}) to compute 3D joints' positions based on the previously estimated configuration.
Then, assuming known camera intrinsics and estimated 2D keypoints, we establish 2D-3D correspondences and apply the Efficient Perspective-n-Point algorithm (EPnP) \cite{epnp}.
%
Any keypoint detector trained to follow the specifications of the robot's coordinate frames can be used to determine the keypoints \cite{dream, lu2023markerless}.
If we assume that the keypoints are perfectly detected, the pose and configuration estimated in this manner are tightly coupled, i.e., the error in the estimated pose is completely determined by the error in the estimated configuration.
To alleviate this effect, we approximate the translational component of ${}_{b}^{c}\mathbf{T}$ by using the kinematic model of the robot and the estimated keypoints, information that we already assume is available. More specifically, we compute the scaling factor $\lambda \in \mathbb{R}$ as follows:
\begin{equation}
\lambda = \frac{\| L_{ij} \|}{\| \tilde{\mathbf{p}}_{i} - \tilde{\mathbf{p}}_{j} \|},
\label{eq:scale_factor}
\end{equation}
where uncalibrated 2D keypoints $\tilde{\mathbf{p}}_{i}, \tilde{\mathbf{p}}_{j}$ bound the chosen link $L_{ij}$ of known length by representing the positions of its child and parent coordinate frames $i,j$.
The initial translational component is defined by the scale factor, camera intrinsics, and a reference keypoint $\mathbf{p}_{ref}$ chosen to represent the reference frame of the robot with respect to the camera.
For simplicity, we choose the base frame of the robot $\mathbf{p}_{ref} = \mathbf{p}_{base}$ and use it to compute (\ref{eq:scale_factor}) together with the keypoint of the neighboring joint.
%
%
%
%

\begin{figure*}[h]
\vspace{0.1cm}
\vspace*{5pt}
\centering
    \begin{subfigure}[b]{0.17\textwidth}
        \centering
        \includegraphics[width=\textwidth, height=2.2cm]{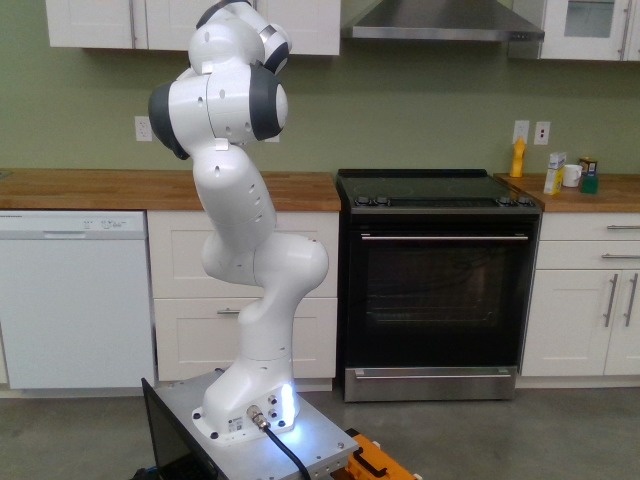}
        \vspace{-3.5mm}
    \end{subfigure}
    \hspace{-0.01\linewidth}
    \begin{subfigure}[b]{0.17\textwidth}
        \centering
        \includegraphics[width=\textwidth, height=2.2cm]{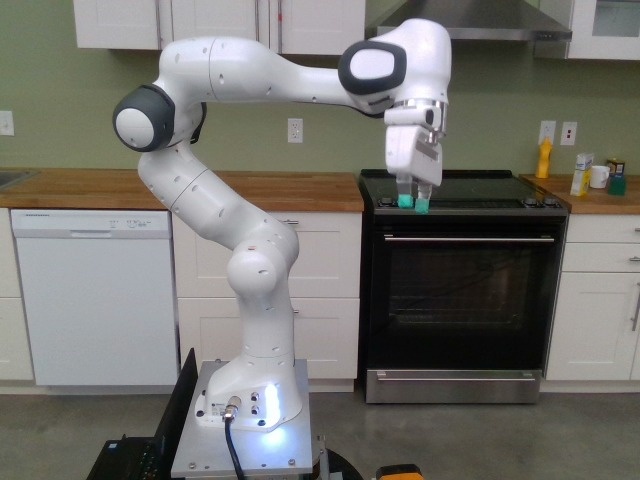}
        \vspace{-3.5mm}
    \end{subfigure}
    \hspace{-0.01\linewidth}
    \begin{subfigure}[b]{0.17\textwidth}
        \centering
        \includegraphics[width=\textwidth, height=2.2cm]{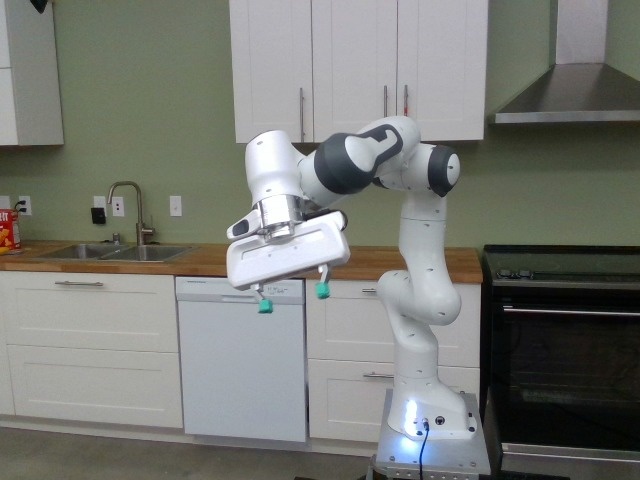}
        \vspace{-3.5mm}
    \end{subfigure}
    \hspace{-0.01\linewidth}
    \begin{subfigure}[b]{0.17\textwidth}
        \centering
        \includegraphics[width=\textwidth, height=2.2cm]{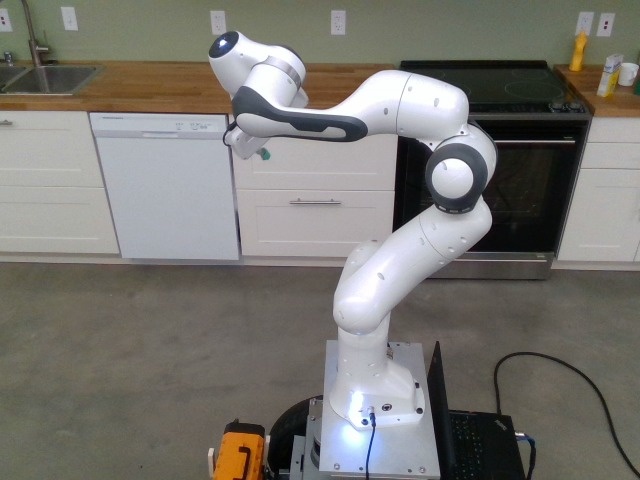}
        \vspace{-3.5mm}
    \end{subfigure}
    \vfill
    \begin{subfigure}[b]{0.17\textwidth}
        \centering
        \includegraphics[width=\textwidth, height=2.2cm]{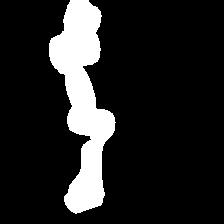}
        \vspace{-3.5mm}
    \end{subfigure}
    \hspace{-0.01\linewidth}
    \begin{subfigure}[b]{0.17\textwidth}
        \centering
        \includegraphics[width=\textwidth, height=2.2cm]{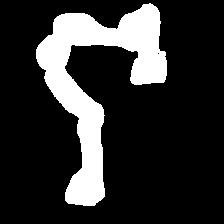}
        \vspace{-3.5mm}
    \end{subfigure}
    \hspace{-0.01\linewidth}
    \begin{subfigure}[b]{0.17\textwidth}
        \centering
        \includegraphics[width=\textwidth, height=2.2cm]{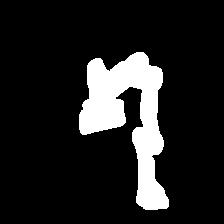}
        \vspace{-3.5mm}
    \end{subfigure}
    \hspace{-0.01\linewidth}
    \begin{subfigure}[b]{0.17\textwidth}
        \centering
        \includegraphics[width=\textwidth, height=2.2cm]{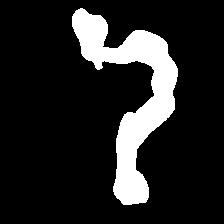}
        \vspace{-3.5mm}
    \end{subfigure}
    \vfill
    \begin{subfigure}[b]{0.17\textwidth}
        \centering
        \includegraphics[width=\textwidth, height=2.2cm]{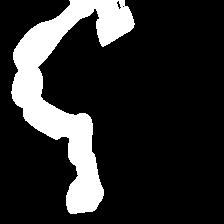}
        \vspace{-3.5mm}
    \end{subfigure}
    \hspace{-0.01\linewidth}
    \begin{subfigure}[b]{0.17\textwidth}
        \centering
        \includegraphics[width=\textwidth, height=2.2cm]{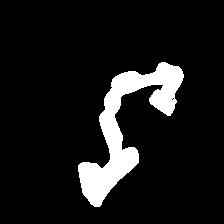}
        \vspace{-3.5mm}
    \end{subfigure}
    \hspace{-0.01\linewidth}
    \begin{subfigure}[b]{0.17\textwidth}
        \centering
        \includegraphics[width=\textwidth, height=2.2cm]{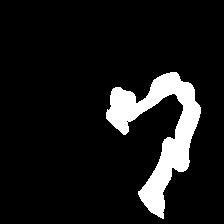}
        \vspace{-3.5mm}
    \end{subfigure}
    \hspace{-0.01\linewidth}
    \begin{subfigure}[b]{0.17\textwidth}
        \centering
        \includegraphics[width=\textwidth, height=2.2cm]{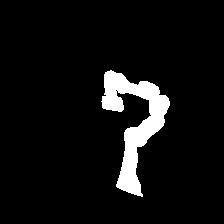}
        \vspace{-3.5mm}
    \end{subfigure}
    \vfill
    \begin{subfigure}[b]{0.17\textwidth}
        \centering
        \includegraphics[width=\textwidth]{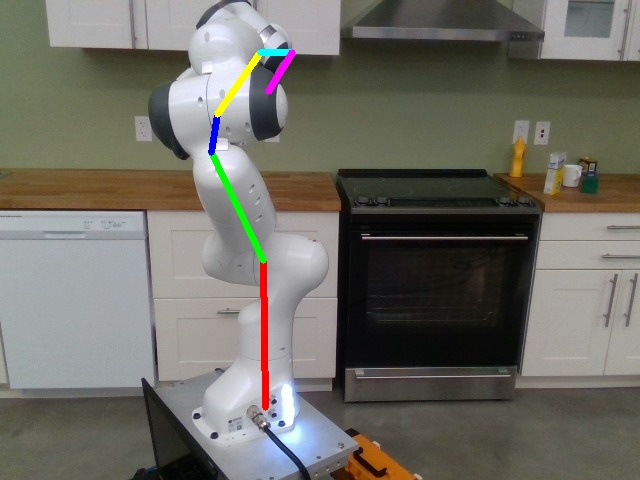}
    \end{subfigure}
    \hspace{-0.01\linewidth}
    \begin{subfigure}[b]{0.17\textwidth}
        \centering
        \includegraphics[width=\textwidth]{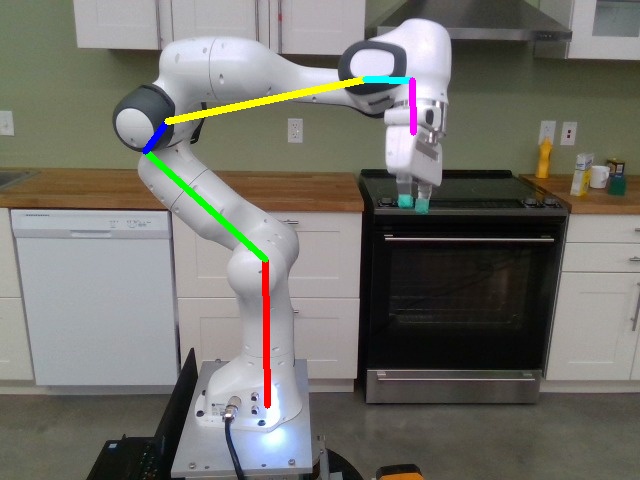}
    \end{subfigure}
    \hspace{-0.01\linewidth}
    \begin{subfigure}[b]{0.17\textwidth}
        \centering
        \includegraphics[width=\textwidth]{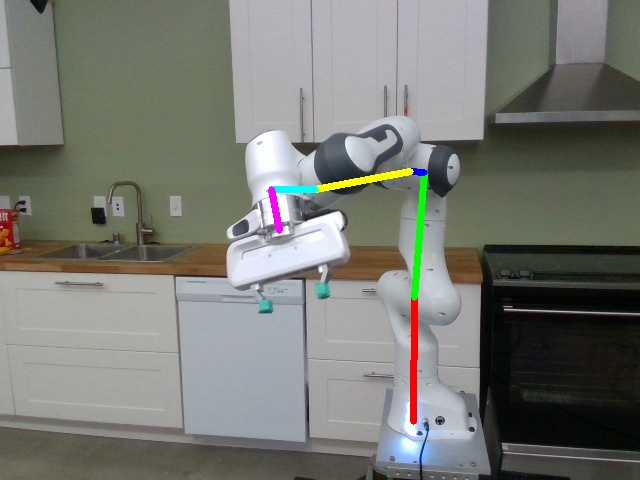}
    \end{subfigure}
    \hspace{-0.01\linewidth}
    \begin{subfigure}[b]{0.17\textwidth}
        \centering
        \includegraphics[width=\textwidth]{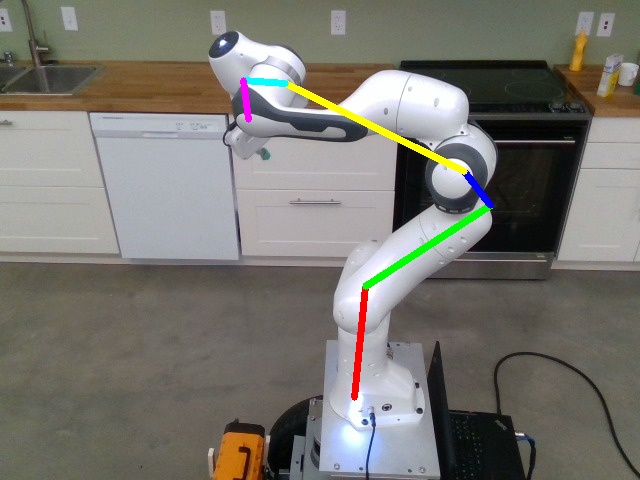}
    \end{subfigure}
    \caption{Qualitative results for pose and configuration estimation; input image (first row), segmented input image (second row), rendered silhouette of an initial estimate (third row) and a projection of the skeleton reflecting the final estimates (last row).}
    \label{fig:qualitative_results}
\end{figure*}

\subsection{Refinement module}
\label{subsec:refinement}
The refinement module uses the initial estimates of robot pose and configuration generated by GIM and estimates their updates based on a rendered silhouette image of the robot defined by the initial estimates.
Given that robotic manipulator consists of rigid links connected in a kinematic chain, each rigid link is visually described by a separate mesh $\mathcal{M}_i$ consisting of a set of vertices ${}^i\mathcal{V} = \{\nu_0, \nu_1, ... ,\nu_n\}$ defined in the local coordinate frame of the $i$-th link.
To render an image, the meshes must be connected to form a continuous chain by transforming each vertex ${}^i\nu \in \mathbb{R}^3$ via respective link-to-base rigid transformation:
\begin{equation}
{}^b\tilde{\nu} = {}_i^b\mathbf{T}(\mathbf{\Theta}) {}^i\tilde{\nu},
\label{eq:mesh_transform}
\end{equation}
where ${}_i^b\mathbf{T}(\mathbf{\Theta}) \in \mathit{SE}(3)$, parametrized by robot configuration, is obtained from forward kinematics
and transforms the vertices into the base coordinate system of the robot.
%
To generate a silhouette image $\mathbb{S}$ representing the initial estimates of the camera-to-robot pose ${}_{b}^{c}\mathbf{T}_{init}$ and configuration $\mathbf{\Theta}_{init}$, points from the robot mesh $\mathcal{M}$ are first sampled uniformly.
These points are then transformed into the camera frame using the estimated pose ${}_{b}^{c}\mathbf{T}$, and projected onto the image plane:
\begin{equation}
\mathbb{S} = \pi(\mathcal{M}, \,{}_b^c\mathbf{T}, \, \mathbf{K}),
\label{eq:generate_silhouette_from_mesh}
\end{equation}
where $\pi(\cdot)$ is the projection operator, while $\mathbf{K}$ represents the camera intrinsics.

Next, an input RGB image is converted into a silhouette image, concatenated with the rendered silhouette and fed to a deep neural network that outputs an update $\Delta \mathbf{\Theta}, \Delta \mathbf{T}$. The configuration is updated by summation:
\begin{equation}
\hat{\mathbf{\Theta}} = \mathbf{\Theta}_{init} + \Delta \mathbf{\Theta},
\label{eq:q_update}
\end{equation}
whereas the update of the pose $\Delta \mathbf{T} = \{\Delta \mathbf{R}, \Delta \lambda\}$ is decomposed into multiplicative updates of the rotation matrix and the scaling factor:
\begin{equation}
\begin{aligned}
{}_{b}^{c} \hat{\mathbf{R}} = \Delta \mathbf{R} \: {}_{b}^{c}\mathbf{R}_{init} \\
\hat{\lambda} = \Delta \lambda \, \lambda_{init}.
\end{aligned}
\label{eq:pose_update}
\end{equation}
The relative pose ${}_{b}^{c}\hat{\mathbf{T}}$ is defined by the updated rotation matrix ${}_{b}^{c} \hat{\mathbf{R}}$ parametrized according to \cite{zhou_continuity}, and the translation $\hat{\mathbf{t}}=\hat{\lambda} \mathbf{K^{-1}} \mathbf{p}_{base}$ computed by back-projecting the 2D detection of the robot's reference frame $\mathbf{p}_{base}$. Thus, the estimated rotation is decoupled from the translation, which in turn is decomposed into respective keypoint and distance estimates.
It is known that estimating the translation in this way is simpler than direct regression \cite{posecnn}.

The model is trained in a supervised manner to predict the pose and configuration updates.
Before calculating the configuration loss, we project the predicted configuration via basic trigonometric functions to account for angular rotation, since the range of joint motion is in the $[0, 2\pi]$ interval.
We therefore compute $\hat{\mathbf{A}} = [ \sin{(\hat{\mathbf{\Theta}})} \oplus \cos{(\hat{\mathbf{\Theta}})} ]$, where $\hat{\mathbf{A}} \in \mathbb{R}^{2n}$ is obtained by concatenation $\oplus$ of $\sin(\cdot)$ and $\cos(\cdot)$, which act on the configuration element-wise. The configuration loss is then:
\begin{equation}
\mathcal{L}_c = \lvert \hat{\mathbf{A}} - \mathbf{A} \rvert,
\label{eq:q_loss}
\end{equation}
where $\lvert \; \cdot \; \rvert$ is the $L_{1}$ norm.
Furthermore, we compute the pose loss by comparing the ground-truth and estimated 3D joint locations defined in the camera frame and obtained by applying the estimated and ground truth pose transformations, respectively:
\begin{equation}
\mathcal{L}_p = \lvert (\hat{\mathbf{R}}, \mathbf{t}) \mathbf{P} - (\mathbf{R, t})\mathbf{P} \rvert + \lvert (\mathbf{R}, \hat{\mathbf{t}}) \mathbf{P} - (\mathbf{R, t})\mathbf{P} \rvert,
\label{eq:pose_loss}
\end{equation}
where $\mathbf{P}$ is computed using forward kinematics and $(\mathbf{R}, \mathbf{t})$ is the ground-truth pose.
Following \cite{robopose}, we use the two different terms to separate the influence of each estimate $\hat{\mathbf{R}}, \hat{\mathbf{t}}$ on the estimated pose, although we only use a sparse set of 3D joint locations compared to a random sample of pose anchors.
The final loss is a linear combination of (\ref{eq:q_loss}) and (\ref{eq:pose_loss}). 
Instead of individual loss weighting, we normalize the gradients w.r.t. output layers to a unit norm to encourage equal contribution of different losses in the shared latent space.
%
%

%
To summarize, instead of feeding the refinement module with random perturbations of the ground truth data as initial estimates during training, we use a trainable geometric initialization module and employ the same setup at test time.
Consequently, the initialization of the refinement module will be closer to a true solution, resulting in a lower running time and more accurate estimates.
%
%
Equally important, we employ a fast rendering procedure that avoids rasterization and shading to generate silhouette images of the robot, which are used to train the refiner.
%
%


\section{EXPERIMENTAL RESULTS}
\label{sec:experiments}
In this section, we present our experimental results and provide ablations using both real and synthetic data.
Specifically, we investigate how the GISR's initialization and refinement modules contribute to the robot pose and configuration estimates, and we examine their impact on the training and running time.
We also analyze how the predictions of the refinement module depend on the availability of varying information from the geometric initialization module and different training data sizes.
Further, we examine the effectiveness of a simple, yet important feature of our method - learning from fast-to-generate silhouette images instead of rendering RGB images via off-the-shelf rendering pipelines.
Finally, we compare GISR with existing state-of-the-art algorithms.

\subsection{Dataset and evaluation metrics}
In all experiments, we used a publicly available Panda-3Cam dataset presented in \cite{dream}.
The dataset contains both real and synthetic images showing a 7-DoF Franka's Panda robotic arm in different configurations and camera-to-robot poses.
%
The synthetic part of the dataset consists of $100$k images, while the real part consists of $50$k images and is divided into four parts; Azure (AK), Kinect (XK), RealSense (RS), and ORB datasets captured using three different cameras. 
AK, XK, and RS subsets of the dataset are used exclusively as test sets, i.e., we never fine-tune on any of them. ORB is used as a training set for ablation experiments or the segmentation model, and is therefore never used as a test set.
%
%
%

Robot pose and configuration estimates are evaluated using the average Euclidean distance (ADD) metric:
\begin{equation}
\text{ADD} = \frac{1}{n} \sum_{i=1}^{n} \| {}_{b}^{c}\mathbf{T} \mathbf{p}_{i} - {}_{b}^{c}\hat{\mathbf{T}} \hat{\mathbf{p}}_{i} \|,
\label{eq:add}
\end{equation}
where ${}_{b}^{c}\mathbf{T}$ and $\mathbf{p}_{i}$ denote the ground-truth camera-to-robot pose and the 3D robot joint keypoints, respectively.
The pose and configuration estimates are evaluated together via ADD, since $\hat{\mathbf{p}}_{i}$ is determined by the estimated configuration and computed using forward kinematics.
We also report the area-under-the-curve (AUC) as an aggregate performance measure, using a $0.1$m threshold value to be consistent with previous work \cite{dream, lu2023markerless, robopose}.

\begin{table}[t]
  \centering
  \begin{tabular}{ccccc}
      \toprule
      \multirow{3}{*}{Module} & \multirow{3}{*}{Running time (s)} & \multicolumn{3}{c}{ADD (m) \greendown} \\     
      \cmidrule(l{6pt}r{6pt}){3-5}
      & & AK & RS & XK \\
      \midrule
      \textit{init only} & 0.034 & 0.159 & 0.227 & 0.238 \\
      \textit{refine only} & 0.049 & 0.061 & 0.107 & 0.073 \\ 
      \textit{init \& refine} & 0.049* & 0.055 & 0.076 & \bftab 0.050 \\ 
      \textit{init \& refine$^{\dagger}$} & 0.049* & \bftab 0.050 & \bftab 0.064 & 0.052 \\ 
      \bottomrule
  \end{tabular}
  \caption{Module-wise ablation using ADD for evaluation on Panda-3Cam datasets. The best results are obtained when the RM is trained on outputs generated by the GIM ($\dagger$). Note: the total running time (*) assumes running the two processes in parallel.}
  \label{tab:init_vs_refine_vs_init+refine}
\end{table}

\begin{figure}[t]
\vspace*{3pt}
\centering
\includegraphics[width=0.4\textwidth]{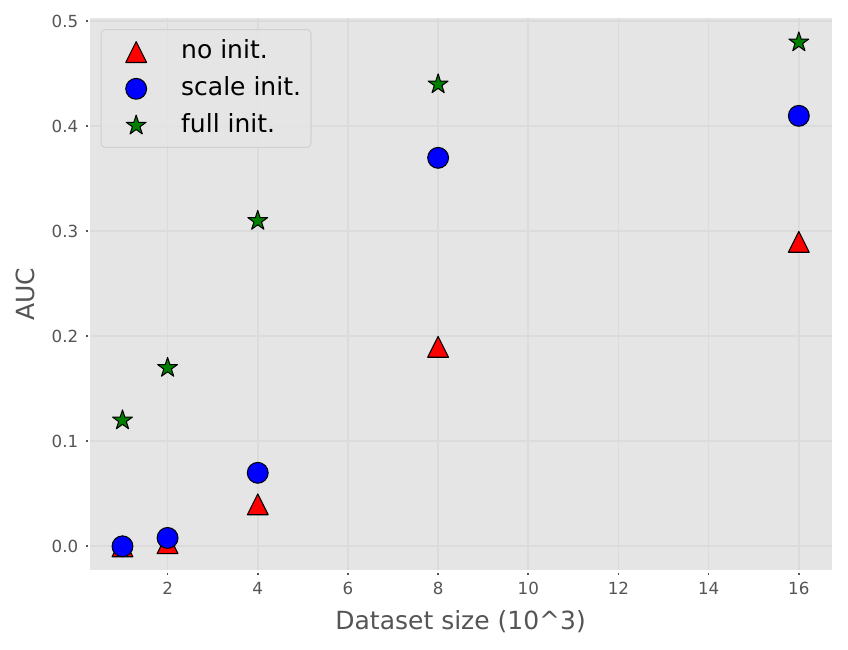}
    \caption{AUC score as a function of training data size for different initialization schemes; (\alignedtriangle) no prior information, (\alignedbullet) initializing scale using 2D keypoints and known robot DH parameters, and (\alignedstar) initializing configuration, scale, and rotation, which amounts to full use of the GIM.}
\label{fig:auc_vs_data}
\end{figure}

\subsection{Training and implementation details}
\label{subsec:impl_details}
%
GISR consists of two trainable modules; geometric initialization (GIM) and refinement (RM) modules.
%
First, we train a shallow, three-layer MLP that performs distance matrix regression as a part of the GIM. 
The rest of this model (i.e., cMDS and IK layers) is differentiable and parameter-free, so that the model comprises only $350$k trainable parameters.
%
Next, we train the RM with initializations generated by the frozen GIM, but whose dropout remains active, making the
outputs of the GIM non-deterministic, which improves the generalization capability of the refiner by mimicking test time.
For the RM, we use a ConvNext-Tiny backbone initialized with ImageNet weights \cite{convnets_2020s}, and train it on 100k synthetic examples to compare it with existing state-of-the-art methods or different sets of real data for other experiments.
The input images are downsampled from a resolution of $640 \times 480$ to $224 \times 224$ to speed up training and inference.
The number of refinement iterations is incremented during training upon reaching $25\%$ and $50\%$ of the total training iterations.
Both models are trained using a batch size of $32$, linear warm-up phase for the first $5\%$ of training iterations, and the Adam optimizer.
For experiments on silhouette-based refinement, input images are segmented at runtime by a pre-trained deeplabv3 \cite{deeplabv3}, which is fine-tuned on $30$k samples of the Panda-ORB by using the silhouette images rendered as described in Section \ref{subsec:refinement} for supervision.
The entire pipeline is implemented to fully utilize GPUs and batching \cite{pytorch3d}, without using \,\emph{for} loops (e.g., for mesh related transformations).
Since training a keypoint detector is beyond the scope of this paper, the 2D keypoint annotations are randomly perturbed by Gaussian noise with a mean of zero and a variance of $30$ pixels.
%
%
%

\subsection{Initialization vs. refinement}
In general, GIM and RM can be used individually, as both are able to predict the camera-to-robot pose and configuration. 
However, GISR uses both modules complementarily to achieve superior results, as shown in Table \ref{tab:init_vs_refine_vs_init+refine}.
To test this idea, we trained GIM and RM both separately and together on $16$k real samples from Panda-ORB.
The GIM predicts a geometric description of the robot, which is used to recover the configuration, while EPnP and forward kinematics are used to compute the 6D pose estimate (\emph{init only}).
Although the error in the order of $20$ cm disqualifies it as a standalone solution, its computational efficiency makes it a suitable initialization method.
On the other hand, rendering-based RM trained using random perturbations of the pose and configuration labels (\emph{refine only}) performs better as a standalone solution, but is also more computationally intensive due to its dense and iterative nature.
%
%
However, when used together (\emph{init \& refine}), the error decreases by $24\%$ on average.
In this setup, we trained RM as in (\emph{refine only}), and used GIM for initialization only at test time.
This shows that GIM can be used out-of-the-box to improve the performance of a pre-trained RM.
The best performance is achieved when the RM is trained based on the results generated by the GIM and the same setting is used at test time (\emph{init \& refine$^{\dagger}$}). This is due to two related reasons.
First, during training, the RM can learn to adapt to the output distribution of the GIM and make larger corrections to the predicted pose and configuration for samples where the GIM produces larger errors, as opposed to training using random ground-truth perturbations where recurrent or similar states are perturbed by different error magnitudes.
%
%
%
Second, the same initialization procedure is used at test time, which serves as a geometric prior and reduces the discrepancy that occurs in the (\emph{refine only}) and (\emph{init \& refine}) setups, where the initialization strategies used at train and test times differ, simply due to the lack of information at test time.
%
\begin{table}[t]
\vspace*{8pt}
  \centering
  \begin{tabular}{cccccc}
      \toprule
      Rendering type & $\mathbf{\Theta}_{gt}$ & AK & RS & XK \\
      \midrule
      RGB & \xmark & 40.12 & 46.05 & 44.07 \\
      Silhouette$^{\dagger}$ & \xmark & 49.64 & 57.93 & 43.13 \\
      Silhouette & \xmark & \bftab 55.48 & \bftab 63.82 & \bftab 53.79 \\
      \midrule
      RGB & \cmark & 53.69 & \bftab 53.56 & \bftab 72.92 \\
      Silhouette & \cmark & \bftab 64.21 & 52.61 & 67.70 \\
      \bottomrule
  \end{tabular}
  \caption{Comparison of RGB- and silhouette-based refinement models via AUC score, evaluated using both real ($\dagger$) and ideal segmentation. Lower part of the table displays results for pose estimation on occluded samples, using known configurations ($\mathbf{\Theta}_{gt}$).}
  \label{tab:rgb_vs_silhouette}
\end{table}
\begin{table}[t]
  \centering
  \begin{tabular}{cccccc}
      \toprule
      Robot & \#DoF & synth. & MAE$(\mathbf{\Theta})$ [deg.] \greendown & ADD$_{ee}$ \greendown & AUC \blueup \\ 
      \midrule
      Panda & $7$ & \xmark & 5.512 & 0.049 & 0.719 \\
      UR5 &  $6$ & \cmark & 5.754 & 0.042 & 0.766 \\
      \bottomrule
  \end{tabular}
  \caption{Performance of the silhouette refinement model on robots with different geometries. The models are trained on 50k synthetic images and tested on 15k (synth. or real) images.}
  \label{tab:diff_robots}
\end{table}
\begin{table*}[t]
\vspace*{8pt}
  \centering
  \begin{tabular}{ccccccccccc}
    \toprule
    \multirow{4}{*}{Method} & \multirow{4}{*}{$\mathbf{\Theta}_{gt}$} & \multirow{4}{*}{Running time (s)} & \multicolumn{2}{c}{Panda-3Cam-AK} & \multicolumn{2}{c}{Panda-3Cam-XK} & \multicolumn{2}{c}{Panda-3Cam-RS} & \multicolumn{2}{c}{all} \\
    \cmidrule(l{6pt}r{6pt}){4-5} \cmidrule(l{6pt}r{6pt}){6-7} \cmidrule(l{6pt}r{6pt}){8-9} \cmidrule(l{6pt}r{6pt}){10-11}
    & & & \multicolumn{2}{c}{\#test\_images = 6394} & \multicolumn{2}{c}{\#test\_images = 4966} & \multicolumn{2}{c}{\#test\_images = 5944} & \multicolumn{2}{c}{\#test\_images = 17304} \\
    \cmidrule(lr){4-5} \cmidrule(lr){6-7} \cmidrule(lr){8-9} \cmidrule(lr){10-11}
    & & & AUC \blueup & ADD (m) \greendown & AUC \blueup & ADD (m) \greendown & AUC \blueup & ADD (m) \greendown & ADD (m) \greendown & MAE$(\mathbf{\Theta})$ \greendown \\ 
    \midrule
    DREAM-F & \cmark & 0.042 & 68.912 & 11.413 & 24.359 & 491.911 & 76.130 & 2.077 & 113.029 & - \\
    DREAM-Q & \cmark & \bftab 0.029 & 52.382 & 78.089 & 37.471 & 54.178 & 77.984 & 0.027 & 59.284 & - \\
    DREAM-H & \cmark & 0.033 & 60.520 & 0.056 & 64.005 & 7.382 & 78.825 & 0.024 & 17.477 & - \\
    DistGeo & \xmark & 0.002 & - & - & - & - & - & - & - & 11.494 \\
    CtRNet & \cmark & 0.033 & \bftab 89.928 & \bftab 0.013 & \bftab 79.465 & \bftab 0.032 & \bftab 90.789 & \bftab 0.010 & \bftab 0.018 & - \\
    \midrule
    RoboPose & \xmark & 1 & 70.374 & 0.033 & \bftab 77.617 & \bftab 0.024 & 74.315 & 0.027 & 0.028 & 5.129 \\
    RoboPose$^{\dagger}$ & \xmark & 0.298 & 63.138 & 0.042 & 27.201 & 0.080 & 43.561 & 0.058 & 0.060 & 11.752 \\
    GISR & \xmark &\bftab 0.049 & \bftab 80.613 & \bftab 0.022 & 73.941 & 0.037 & \bftab 79.324 & \bftab 0.021 & \bftab 0.026 & \bftab 4.808\\
    \bottomrule
  \end{tabular}
  \caption{Comparison with state-of-the-art methods. Evaluation protocol from existing work is followed by using an AUC with a $0.1$m threshold in addition to ADD. Additionally, mean absolute error over predicted configurations (deg.) is reported.}
  \label{tab:ours_vs_sota}
\end{table*}
\subsection{Varying initial information}
In the following, we investigate how varying the initial information about the observed robot by excluding different components of the GIM affects the final performance and scalability of GISR with real training data, especially for small to medium data sets.
The results are shown in Fig.~\ref{fig:auc_vs_data}.
%
Different model variants are created by incrementally adding new information to the initial estimate, trained using different amounts of data and evaluated using the AUC score.
%
The first variant (\emph{no init.}) is a pure refinement model that is trained using random perturbations of the ground truth to simulate a scenario without prior information, and evaluated using a fixed initialization; unit scaling factor, rotation that aligns the camera frame to the robot base frame, and a configuration that corresponds to the middle of the joint limits.
This model serves a baseline.
In the second variant (\emph{scale init.}), 2D joint keypoints and known robot DH parameters are included, enabling initialization of the scale factor used at both train and test time.
In the last variant (\emph{full init.}), we fully utilize the GIM by including robot configuration estimation and EPnP-based rotation estimation, and use the same strategy at both train and test times.
All variants scale similarly, except for small amounts of training data where (\emph{no init.}) scales slower.
However, the (\emph{full init.}) variant is clearly the best-performing across all training data sizes.
In addition, the model exhibits greater sensitivity to initial scale than to rotation. This is reflected in a larger performance gain between (\emph{no init.}) and (\emph{scale init.}) compared to that between (\emph{scale init.}) and (\emph{full init.}).


\subsection{Learning from silhouette images}

Existing rendering-based methods for pose estimation of objects \cite{deepim, manhardt2018deep} and robots \cite{robopose} typically use RGB rendering.
In contrast, our refinement model learns to predict the camera-to-robot pose and configuration updates based on silhouette images.
These two approaches are compared. 
In the former, we use the reference and rendered RGB images that reflect the observation and the current estimate, respectively.
The latter corresponds to GISR, by taking the respective silhouette images as input instead, where the reference RGB image is converted into a silhouette via segmentation.
The silhouette image reflecting the current estimate is computed as described in Section~\ref{subsec:refinement}, avoiding the costly steps typically involved in rendering, i.e., rasterization and shading.
As shown in Table \ref{tab:rgb_vs_silhouette}, the model trained on silhouette images outperforms its RGB counterpart.
Moreover, our experiments show that this approach achieves at least a twofold speedup compared to an optimized off-the-shelf RGB renderer \cite{pytorch3d}.
The lower part of the table presents results on images with occlusions. We manually selected 180  images from Panda-3Cam test datasets, where the pose and configuration result in robot self-occlusions. The two methods exhibit similar performance, suggesting that the silhouette approach is as robust as its RGB counterpart.

Assuming a model-based approach like ours, a silhouette image of the robot is fully defined by its configuration and the camera-to-robot pose. Thus, it contains all the necessary information for the model to learn while omitting irrelevant details present in RGB images.
In the context of our task, only a small subset of information in RGB images adds value for generalization (e.g., salient color patterns on the robot). This makes the model more susceptible to overfitting, as it must learn to ignore most of the irrelevant details (e.g., background, lighting conditions).
Similarly, RGB-based methods are more affected by the fidelity gap between input and rendered images due to imperfections in the rendering process, the 3D model of the robot, and variations in background and lighting conditions.
In contrast, learning from silhouette images limits the model to focus only on the most important information, offers faster rendering times, and reduces discrepancies in fidelity. However, this comes at the cost of requiring an additional model.

The last row in the upper part of Table \ref{tab:rgb_vs_silhouette} illustrates how robot segmentation performance affects the estimation of the camera-to-robot pose and configuration.
To measure this, we used our fast silhouette rendering to segment input images by leveraging pose and configuration labels. This simulates the scenario of having a close-to-ideal segmentation model at test time.
The results clearly show that improving the segmentation model alone results in much more accurate pose and configuration estimates.
Notably, this approach does not require human annotations at the pixel level, which are commonly used to train segmentation models. Instead, the model is supervised using simplified silhouette rendering, leveraging the actual robot configuration and camera-to-robot pose, which are less time-consuming and easier to capture.

Finally, to demonstrate the generality of our approach for robot arm pose estimation, we tested it on a synthetic UR5 arm dataset (see Table \ref{tab:diff_robots}), obtaining results that suggest our method does not disproportionately benefit from the specific link geometries of individual robot arms.

\subsection{Comparison to state-of-the-art methods}

Here, we present the evaluation of our method in comparison to existing state-of-the-art methods. The results are shown in Table \ref{tab:ours_vs_sota}.
DREAM \cite{dream} and CtRNet \cite{lu2023markerless} are keypoint-based methods that recover only 6D camera-to-robot pose, for which they require ground-truth configuration at test time ($\mathbf{\Theta}_{gt}$).
DistGeo \cite{bilic_maric_ifac_2023} is a lightweight distance-geometric method that can be applied on top of a keypoint detector to recover only the configuration.
In contrast, RoboPose \cite{robopose} is an RGB rendering-based method trained to recover both the robot pose and configuration, similar to our refinement model.
Our method is hybrid in that it leverages sparse information for initialization, followed by dense silhouette-based refinement to arrive at a final solution.

As reported by authors, DREAM and RoboPose were trained on $100$k synthetic samples from \cite{dream}. To ensure a fair comparison, we trained GISR using only this data. CtRNet was pre-trained on this data but also fine-tuned on the real parts of the dataset, while DistGeo was trained on $10$k samples as it is a tiny model that does not benefit from additional data \cite{bilic_maric_ifac_2023}.
DREAM-H, the best-performing among the DREAM variants, is outperformed by GISR on each of the individual datasets.
Although CtRNet outperforms GISR by $0.008$ ADD, it still relies on ground-truth configurations at test time.

The lower part of Table \ref{tab:ours_vs_sota} is dedicated to methods designed to predict both the pose and configuration.
We compare GISR with two RoboPose variants; the best-performing variant and the one using the same number of iterations as our method ($\dagger$).
GISR achieves slightly better results than the best-performing RoboPose model while being $20\times$ faster.
When we restrict RoboPose to the same number of iterations (3) as GISR, GISR significantly outperforms it while still running $6\times$ faster.
%
In terms of overall performance, our method is competitive with keypoint-based methods that use ground-truth robot configuration.
%



\section{Conclusion}
In this paper, we present GISR, a novel approach to jointly estimate camera-to-robot pose and configuration by utilizing both sparse and dense information.
Our architecture uses distinct geometric initialization and refinement modules to accurately estimate both quantities, which enables improving the final performance by fine-tuning specific subtasks.
The proposed silhouette-based refinement can be executed with low computation times, making it suitable for online estimation and dynamic scenarios with mobile robots.
However, the use of multiple modules also increases the overall complexity and memory requirements compared to end-to-end approaches. 
For future work, we consider extending this approach to unknown robots, i.e., to discover their kinematic model.

\bibliography{ieeetran}
\bibliographystyle{IEEEtran}

\end{document}